\documentclass[letterpaper, 10 pt, conference]{ieeeconf}  

\IEEEoverridecommandlockouts                              

\usepackage{amsmath} 
\usepackage{amsfonts} 
\usepackage{todonotes}
\usepackage{wrapfig}
\usepackage{cite}
\usepackage{hyperref}

\title{\LARGE \bf
Train What You Know --- Precise Pick-and-Place with Transporter Networks
}

\author{Gergely S\'{o}ti$^{1,2}$, Xi Huang$^2$, Christian Wurll$^1$, Björn Hein$^{1,2}$
\thanks{This research is being conducted as part of the KI5GRob project funded by the German Federal Ministry of Education and Research (BMBF) under project number 13FH579KX9.}
\thanks{The authors are with $^1$Robotics and Autonomous Systems, Institute of Applied Research, Karlsruhe University of Applied Sciences, 76133 Karlsruhe, Germany, and $^2$Institute for Anthropomatics and Robotics, Karlsruhe Institute of Technology, 76131 Karlsruhe, Germany \newline{\tt\small gergely.soti@h-ka.de}}%
}

\DeclareMathOperator*{\argmax}{arg\!\max}

\begin{document}

\maketitle
\thispagestyle{empty}
\pagestyle{empty}

\begin{abstract}

Precise pick-and-place is essential in robotic applications. To this end, we define a novel exact training method and an iterative inference method that improve pick-and-place precision with Transporter Networks \cite{zeng2020transporter}. We conduct a large scale experiment on 8 simulated tasks. A systematic analysis shows, that the proposed modifications have a significant positive effect on model performance. Considering picking and placing independently, our methods achieve up to 60\% lower rotation and translation errors than baselines. For the whole pick-and-place process we observe 50\% lower rotation errors for most tasks with slight improvements in terms of translation errors. Furthermore, we propose architectural changes that retain model performance and reduce computational costs and time. We validate our methods with an interactive teaching procedure on real hardware. Supplementary material will be made available at: 
\url{https://gergely-soti.github.io/p3}
\end{abstract}

\section{Introduction}



Although learning grasping as a standalone module has been researched for many years, the whole pick-and-place pipeline answers the question: What to do after picking up an object? In terms of industrial applications, assembly or packaging are leading examples but pick-and-place is a fundamental skill in the growing area of service robotics. This wide application range comes with a variety of difficult problems regarding generality, flexibility, robustness, and reliability. To solve these, recent advancements in deep learning provide various powerful tools. With computer vision, as one of the main fields in AI research, vision-based robotic manipulation gained a still lasting momentum. Researchers aim to create more general and versatile methods that learn faster and better. Goals, among others, are handling unknown objects, achieving higher precision, autonomously solving complex and sequential manipulation tasks, improving sample efficiency, or even removing the need for labeled data via self-supervised learning.



In this work we propose a novel \textit{exact} training method, an iterative inference method and some architectural changes to Transporter Networks (TN) \cite{zeng2020transporter} to improve their performance on planar pick-and-place tasks. We conduct a large scale experiment, to analyze the effects of these modifications, and evaluate our methods on simulated tasks, achieving up to 60\% lower rotation and translation errors at picking and placing than the baseline, and 50\% lower rotation errors for the whole process. Finally, we demonstrate the effectiveness of the proposed methods on a real world setup using an interactive method to teach the robot to autonomously solve a pick-and-place task from 10 demonstrations.

\begin{figure}[!]
\centering
  \includegraphics[width=0.4\textwidth]{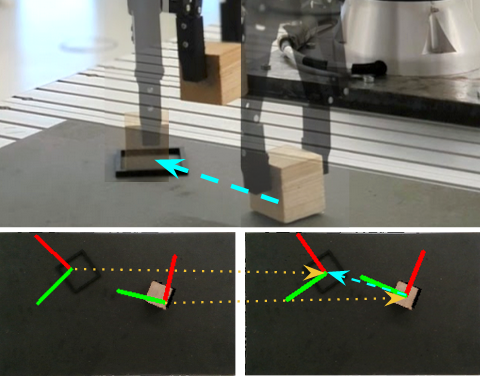}
  \caption{During interactive teaching, the robot proposes pick and place poses, which the user can correct. The corrected poses are incorporated to the training and contribute to improving future proposals.}
  \label{fig:transport-real}
\end{figure}

\section{Related Work}

One approach to solve pick-and-place tasks is to couple object detection or segmentation with 6D pose estimation and successful grasp pose estimation \cite{zhu2014single,zeng2017multi,xiang2017posecnn,wang2019normalized,deng2020self}. These methods however, often require object models and objects need to have a nominal direction, which make these methods unfit for handling unknown or deformable objects. Methods based on keypoint detection or dense descriptors can generalize to object categories and can even learn to manipulate deformable objects, but often require a large amount of object-specific labeled data making their universal application difficult \cite{florence2018dense, manuelli2019kpam, kulkarni2019unsupervised, nagabandi2020deep, liu2020keypose}. End-to-end learning models tend to work better in unstructured environments (logistics, household), as they directly map perception to actions \cite{gualtieri2018pick, wu2019learning, zakka2020form2fit, hundt2020good, berscheid2020self, devin2020self, khansari2020action, song2020grasping, zeng2020transporter}. In this context, the recently proposed Transporter Network architecture \cite{zeng2020transporter} is gaining popularity mainly due to its plausible design and sample efficiency (detailed description in \ref{subsec:transporter-networks}). It has various modifications and extensions that incorporate language policies \cite{shridhar2022cliport}, further improve sample efficiency through complete rotational equivariance \cite{huang2022equivariant}, enable solving complex sequences of different tasks \cite{lim2022multi} or deal with deformable objects via goal conditioning \cite{seita2021learning}.

\section{Method}
We consider the problem of learning planar pick and place actions in form of SE(2) poses from visual observations:
\begin{equation}
    f: \mathbb{R}^{w \times h \times  4} \to SE(2) \times SE(2); \mathbf{o} \to (\mathcal{T}_{pick}, \mathcal{T}_{place})
\end{equation}
with $\mathbf{o} \in \mathbb{R}^{w \times h \times  4}$ an orthographic projection of a pointcloud, an image with $w$ width, $h$ height and 4 channels for RGB-D. $\mathcal{T}_{pick} \in SE(2)$ and $\mathcal{T}_{place}  \in SE(2)$ correspond to endeffector poses. Although this formulation is not only object but also gripper agnostic, we restrict the problem to using a 2-jaw parallel gripper. Note, that such a pick-and-place task can have multiple valid solutions, as there can be multiple valid pick poses and multiple valid corresponding place poses.


In this section, first we describe TNs \cite{zeng2020transporter} followed by our proposed architectural changes. Then we present the different training and inference methods, which we consider as our main contribution. We conclude this section with the methodology for evaluating these.

\subsection{Transporter Networks}
\label{subsec:transporter-networks}
Transporter Networks address the problem of learning manipulation actions from visual observations, and consist of two fully convolutional neural network modules. The attention module learns to predict the gripper's position for successfully grasping an object and the transport module uses this inferred pick position to predict the gripper pose for precise placement. Both modules rely heavily on convolution operations which are applied to and benefit from spatially consistent visual representations, like orthographic projection of pointclouds. Such projection ensures, that planar movement of an object would not change its appearance in the observation in contrast to distortions when using plain camera images. This also makes extensive data augmentation possible, as transforming observations and corresponding poses results in valid data.

The attention module $F_{pick}$ maps the observation $\mathbf{o}$ to a feature map corresponding to probability distribution of successful grasping over pick positions. The correct pick position is obtained by maximizing $F_{pick}(\mathbf{o})$ over possible positions $(u, v)$:
\begin{equation}
\begin{gathered}
    F_{pick}: \mathbb{R}^{w \times h \times  4} \to \mathbb{R}^{w \times h \times  1} \\
    (u_{pick}, v_{pick}) = \argmax_{u, v} F_{pick}(\mathbf{o})
\end{gathered}
\end{equation}
$(u_{pick}, v_{pick})$ compose $\mathcal{T}_{pick}$, since \cite{zeng2020transporter} used a vacuum gripper, thus rotations are not considered at picking. 

The transport module is conditioned on the detected pick pose and learns to correlate visual features of the picked object with visual features of the object's target pose. Through the mapping $F_{query}$, the object's features are learned as a dense embedding of the cropped observation around the pick position denoted by $\mathbf{o}[\mathcal{T}_{pick}]$. Another dense embedding is learned over the whole observation using $F_{key}$, an illustration see Fig. \ref{fig:transport-exact}. The correlation of the query embedding $F_{query}(\mathbf{o}[\mathcal{T}_{pick}])$ and a partial crop around $\mathcal{T}$ from the key embedding $F_{key}(\mathbf{o})[\mathcal{T}]$ correspond the the probability of correct object placement in $\mathcal{T}$ given the pick pose $\mathcal{T}_{pick}$. To account for differently rotated objects, the key embedding is calculated not only for $\mathbf{o}[\mathcal{T}_{pick}]$ but for its discretely rotated copies is also calculated. The correlation of these embeddings and the key embedding determine the place pose:
\begin{equation}
\begin{gathered}
    F_{query}: \mathbb{R}^{n \times c \times c \times  4} \to \mathbb{R}^{n \times c \times c \times  3} \\
    F_{key}: \mathbb{R}^{w \times h \times  4} \to \mathbb{R}^{w \times h \times  3} \\
    rot^\alpha_n: \mathbb{R}^{c \times c \times  d} \to \mathbb{R}^{n\times c \times c \times  d} \\
    \mathbf{q} = F_{query}(rot^\alpha_n(\mathbf{o}[\mathcal{T}_{pick}]))    \\
    (u_{place}, v_{place}, r_{place}) = \argmax_{u, v, r} \mathbf{q} \ast F_{key}(\mathbf{o})
\end{gathered}
\label{eq:crop_rot}
\end{equation}
Both $F_{query}$ and $F_{key}$ map an orthographic projection to an embedding with depth 3. $c$ is the width and height of the cropped image patch. $rot^\alpha_n$ denotes the function, that maps an image patch to its $n$ rotated copies with $\alpha$ rotation step. $d$ refers to the number of channels in the image patch, which is 4 in case of RGB-D data. In TNs $n = 36$ and $\alpha = 10^{\circ}$. $(u_{place}, v_{place}, r_{place})$ compose $\mathcal{T}_{pick}$. $F_{pick}$, $F_{query}$, $F_{key}$ are approximated by 43-layer ResNet hourglass fully convolutional neural networks.

\subsection{Model Specifics}
\subsubsection{Attention for position and rotation}
\begin{figure}[!]
  \includegraphics[width=0.49\textwidth]{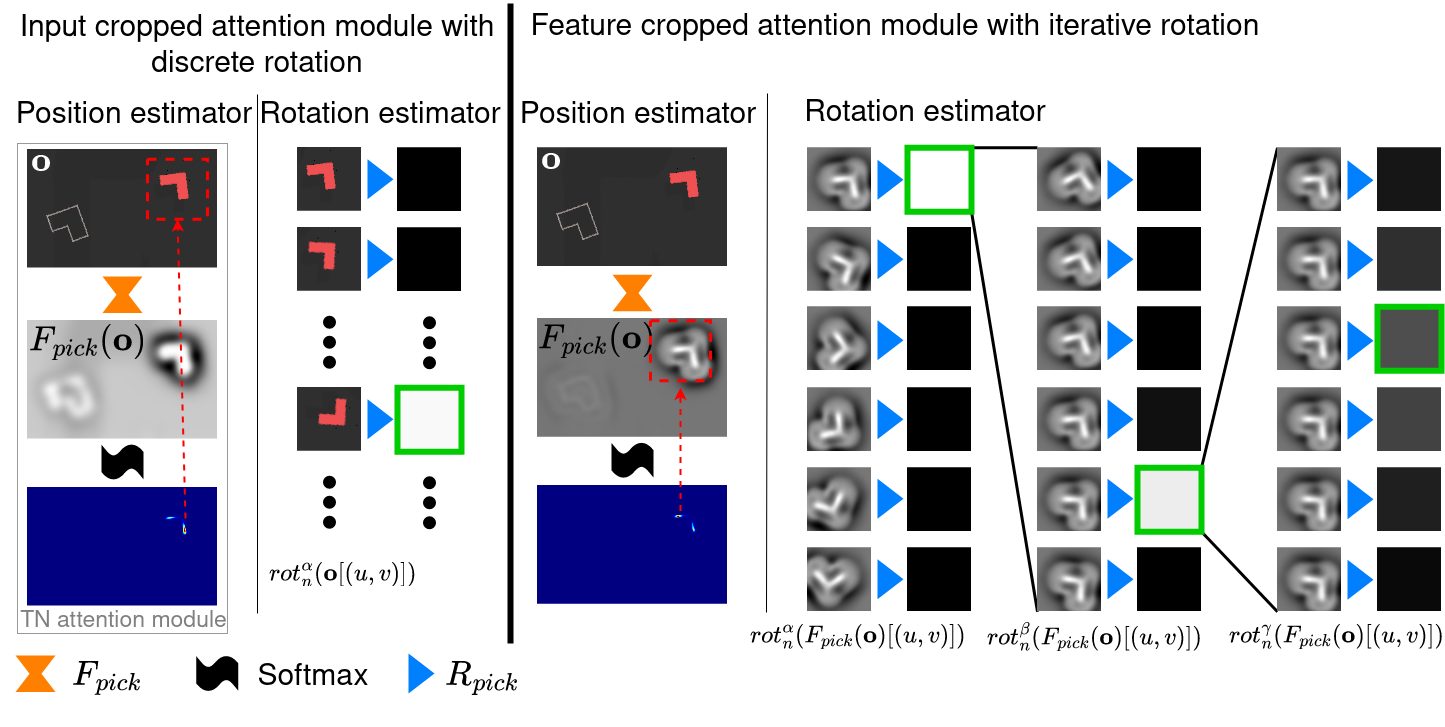}
  \caption{Input cropped attention modules (left) estimate position with a model that is identical to TN's attention module and estimate rotation from an image patch from the observation independently. In case of feature cropping (right), rotation estimation also influences position estimation, as it uses the features from position estimation which allows a rotation error backpropagation through the position estimator network. Discrete rotation estimation (left) rotates image patches into predefined rotation bins (10°), propagates these through the same network as a batch and learns to detect correct rotations using cross entropy loss. Iterative rotation estimation (right) does the same but with less bins and an increasing resolution while using the results from the previous iteration.}
  \label{fig:attention-iterative}
\end{figure}

We extend TNs' attention module to consider rotations at picking. Although the published TN implementation is theoretically able to predict rotations, it requires excessive amount of GPU memory and is thus not feasible.
Similar to \cite{huang2022equivariant}, we decompose the attention module into a position and an rotation estimator. The position estimator $F_{pick}$ is identical to TNs' attention module and computes $(u_{pick}, v_{pick})$. For rotation estimation we use a single convolutional layer $R_{pick}$ with a receptive field as big as its input and a single output neuron, which makes it basically a fully connected layer. It processes a batch of rotated copies of $(u_{pick}, v_{pick})$ centered image patches. We propose two variants (see Fig. \ref{fig:attention-iterative}). In the first variant (input cropped, $r_{pick}^{ic}$), the image patch is taken from the observation $\mathbf{o}$:
\begin{equation}
    r_{pick}^{ic} = \argmax_{} R_{pick}(rot^\alpha_n(\mathbf{o}[(u_{pick}, v_{pick})])).
\end{equation}
In the second variant (feature cropped, $r_{pick}^{fc}$) the image patch is taken from the feature embedding $F_{pick}(\mathbf{o})$:
\begin{equation}
    r_{pick}^{fc} = \argmax_{} R_{pick}(rot^\alpha_n(F_{pick}(\mathbf{o})[(u_{pick}, v_{pick})])).
\end{equation}
During training, a cross-entropy loss is used for maximizing the output for the correctly rotated patch and inhibiting the outputs for the other rotations during training. $(u_{pick}, v_{pick}, r_{pick})$ compose $\mathcal{T}_{pick}$




Rotated copies are calculated using bilinear interpolation, which is differentiable. This leads to the main difference between the two variants and the motivation for the second: the gradients from rotation estimation flow back into the position estimator network. With this, we aim to utilize the capacity of the position estimator network, with the possibility of learning features that are easier to classify for the rotation estimation network.


\subsubsection{Transport}
In TNs' transport module, query and key networks are asymmetrical with regard to rotational equivariance. Whereas $F_{query}$ also processes rotated copies $\mathbf{o}[\mathcal{T}_{pick}]$, thus is not required to learn the rotations themselves, $F_{key}$ only receives $\mathbf{o}$ in its original orientation. This means, that the visual cues of target poses are not learned to be recognized in all orientations, rather only in those that were presented during training.
Although not inherently rotationally invariant, TNs demonstrated that by employing data augmentation, $F_{key}$ can learn to extract suitable features, that correlate to query features of differently rotated $\mathbf{o}[\mathcal{T}_{pick}]$. This motivates to the introduction of a different usage of the $F_{query}$, as in only computing the feature embedding on the non-rotated image patch $\mathbf{o}[\mathcal{T}_{pick}]$ and then rotating the embedding only for calculating the correlations (see Fig. \ref{fig:transport-exact}):
\begin{equation}
\begin{gathered}
    \mathbf{q} = rot^\alpha_n(F_{query}(\mathbf{o}[\mathcal{T}_{pick}]))    \\
    (u_{place}, v_{place}, r_{place}) = \argmax_{u, v, r} \mathbf{q} \ast F_{key}(\mathbf{o})
\end{gathered}
\label{eq:query_rot}
\end{equation}



This leads to reduced computational costs, as the query network only has to be executed once, instead of multiple times for each rotated copy of $\mathbf{o}[\mathcal{T}_{pick}]$ . In the following, we refer to the original transport module implementation as crop rotated transport module, and to our variant as query rotated transport module.

\subsection{Training and Inference Methods}
Finding the correct orientation of the gripper for both picking and placing relies on rotating a partial crop or its feature embedding into its correct orientation, so that the output value of $R_{pick}$ or the correlation calculation at the correct position (eq. \ref{eq:crop_rot} and \ref{eq:query_rot}) is maximized. Intuitively, the more correct this rotation is, the higher are the corresponding output values of $R_{pick}$. This is true at least in some environments of the correct orientation, which is a sufficient condition for defining an iterative rotation estimation method. Essentially, this means that the methods, that estimate rotations are executed successively multiple times with differently parametrized $rot_n^\alpha$, building up on the previous iteration. The first iteration encompasses the whole rotation space. In subsequent iterations, $\alpha$ needs to be decreased, and the new rotated copies need to be centered around the last iteration's result (see Fig. \ref{fig:attention-iterative}). Note, that this only alters how a network is executed, not its architecture. As a consequence, networks trained with discretized rotation representation can be used with iterative rotation estimation at inference time and vice versa. 


Considering training using iterative rotations, leads to the recognition of a fundamental flaw. The problem lies with what we consider as a correct rotation when calculating the loss. With for example 30° in the first, 5° in the second and 0.83° in the third iteration, orientations that are incorrect in the last iteration would be considered correct in the second and first iterations. This sends contradicting signals during optimization and could lead to unstable training.
This is also present when training with the discrete rotation formulation, or when using any kind of discretization. In case of a 10° resolution, we have a 5° error margin that we still consider correct, and label solutions that are incorrect as correct. 

\begin{figure}[!]
  \includegraphics[width=0.485\textwidth]{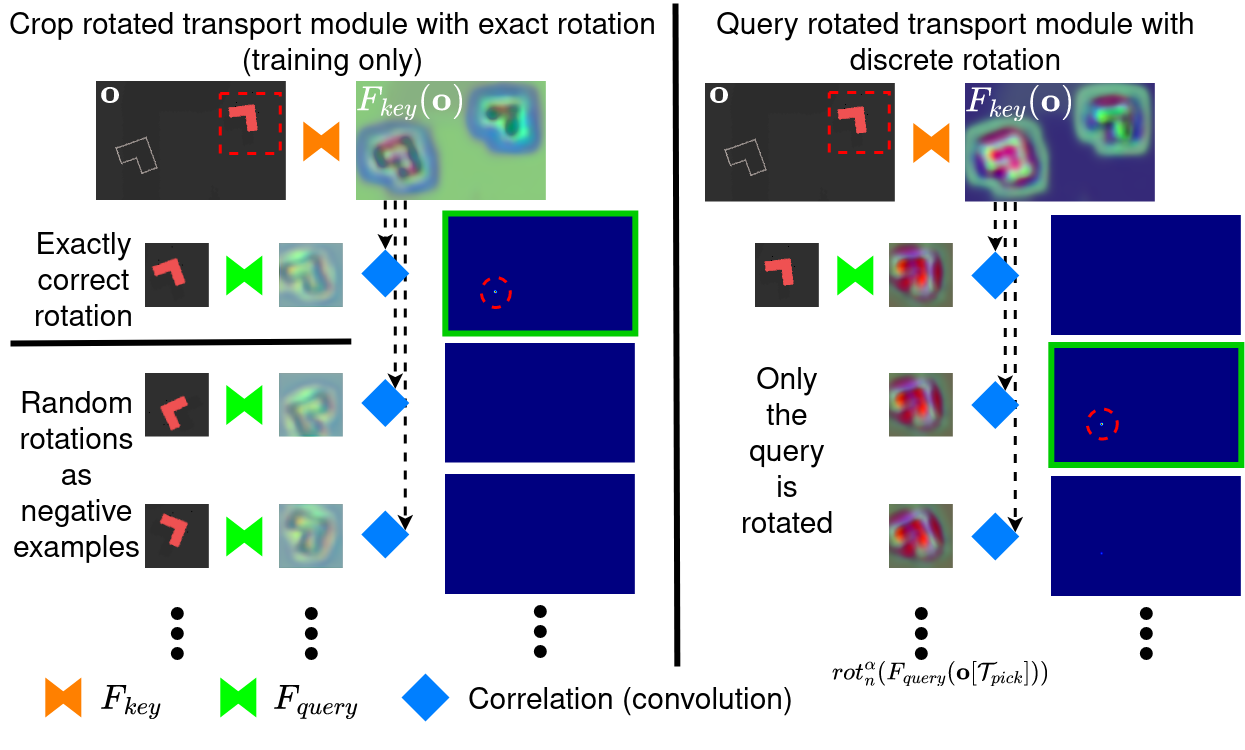}
  \caption{Transport modules estimate translation and rotation simultaneously by correlating a query feature map with a key feature map. Crop rotated transport (left) computes the query feature map for differently rotated image patches from the observation, while query rotated (right) computes the query feature map only once and rotates this feature map before correlation. Discrete rotation estimation (right) uses predefined rotation bins (10°), as opposed to exact rotation (left, training only) which uses one exactly rotated image patch as positive example and multiple randomly rotated versions as negative examples.}
  \label{fig:transport-exact}
\end{figure}

We address this issue by introducing the exact training method (see Fig. \ref{fig:transport-exact}). Instead of rotating the image patches by some pre-defined rotation steps, we rotate them to the exact solution to provide a positive example. This way, we exactly \textit{train what we know}. For negative examples, we simply rotate the image patch by random angles as these result in incorrect orientations with a very high probability. Of course, rotational symmetries could cause unexpected behavior, but only if $\mathcal{T}_{pick}$ is in the rotational symmetry axis, which is rarely the case.

We parametrize the exact training method by the number of negative samples that we randomly generate during training. As for iterative inference, we use the number of iterations, the number of distinct rotations in an iteration and the step scaling parameter, which defines the scaling factor applied to the rotational space after each iteration. 
\subsection{Model Analysis and Selection}
\label{subsec:analysis}
We consider the different attention and transport module variants, as well as the training and inference methods as model hyperparameters, along with the number of demonstrations we use for training and the number of weight update steps during training. We want to analyse and evaluate their influence on model performance in terms of attention and transport errors (see \ref{subsubsec:metrics}).
To achieve this, we define the function $H$, that maps a hyperparameter configuration to the expected value of the given metric. We train multiple models with different hyperparameters and evaluate them. With the obtained data we train a random forest regressor to approximate $H$. The hyperparameters act as features in the model, thus we can measure their importances. The larger the importance of a hyperparameter, the more it influences the correctness of the regression. This means, that the expected value for the metric is dependent on the choice of that hyperparameter, thus it correlates to model performance. We use GINI importance and permutation importance to rank the hyperparameters.

Our main goal is to increase the precision in pick-and-place tasks. For this, first, we rank the hyperparameters based on their influence on model performance, as described above. Then we filter out worst performing value of the most influential hyperparameter. We repeat this, until we arrive at a single hyperparameter combination or until the choice of hyperparameter value does not affect model performance (as per the learned performance prediction model). 



\section{Experiments}
We conduct experiments both in simulation and on real hardware. We perform a large experiment on simulated tasks, and analyse the effects of different hyperparameters on models' performance. Based on the analysis, we select the best performing models and validate them on a real robot.
We train our attention and transport modules independently.
As baseline, we use the hyperparameter configuration, that is closest to the original TN implementation: input cropped attention module and crop rotated transport module. both with discrete training and discrete inference.


Additionally, we investigate whether the improvements are retained when the models are trained and evaluated on a real world robot.
\subsection{Simulated Experiments}
We select block insertion and assembling kits from the Ravens-10 benchmark, to test our methods. We modify these tasks as described below to accommodate for using a 2-jaw parallel gripper. We also include additional objects in block insertion to isolate the models' behavior when dealing with different kinds of object, especially in terms of rotational symmetries.




\subsubsection{Tasks}
\begin{itemize}
    \item \textbf{Block insertion:} picking up a block and placing it into a fixture. We lowered fixture height to avoid collision with the gripper. We use different objects, that can be grouped by their rotational symmetries: L-shape, T-shape (no rotational symmetries); cuboid, long cuboid, double-L-shape (2 rotational symmetries); cube, cross (4 rotational symmetries). Each object has its own task, meaning that a block insertion dataset only contains demonstrations for exactly one object.
    \item \textbf{Assembling kits:} picking five differently shaped objects and placing them onto their corresponding locations on a kitting board. 
    We use cube, long cuboid, T-shape, cross, U-shape, double-L-shape and a parallelogram based prism for training, and cuboid, L-shape, block-ring and a hexagon based prism for testing. In this task, models are required to generalize to unseen objects.
\end{itemize}

Objects for the tasks are spawned in random, non-overlapping configurations.
We implement a scripted oracle for each task, that provides expert demonstrations for training and can calculate errors during testing. 
Expert demonstrations are generated by sampling the possible pick poses, and from the possible object placement poses, also accounting for the gripper's picking offset.
A demonstration of a task consists of completely solving the task. This means one pick-and-place action for block insertion and 5 pick-and-place actions for assembling kits.

 \setlength{\tabcolsep}{3pt}
\begin{table*}[t]
  \centering
  
  \begin{tabular}{l||rr|rr|rr|rr||rr|rr|rr|rr}
Training samples & \multicolumn{2}{c|}{1} & \multicolumn{2}{c|}{10} & \multicolumn{2}{c|}{100} & \multicolumn{2}{c||}{1000} & \multicolumn{2}{c|}{1} & \multicolumn{2}{c|}{10} & \multicolumn{2}{c|}{100} & \multicolumn{2}{c}{1000} \\
\hline
\hline
 & \multicolumn{16}{c}{\textbf{Attention error}} \\
\hline
Tasks & \multicolumn{8}{c||}{Insertion - no rotational symmetry} & \multicolumn{8}{c}{Insertion - 2 rotational symmetries}  \\
\hline
 baseline & \underline{3.02 }&  5.27 & 1.26 & 2.62 & 1.24 & 2.46 & 1.25 & 2.45 & \underline{2.50} & 3.92 & 1.35 & 2.36 & 1.36 & 2.25 & 1.27 & 2.24 \\
 di, it4, i-crop & 3.17 &  \underline{4.50} & 1.03 & \underline{1.42} & 0.54 & \underline{0.98} & \textbf{0.53} & \textbf{0.79} & 2.78 & \underline{3.49} & 0.96 & 1.62 & 0.90 & 1.41 & 0.85 & 1.26 \\
 di, it4, f-crop & 6.45 & 12.25 & \underline{0.83} & 1.81 & \underline{0.50 }& 1.10 & 0.56 & 1.00 & 3.17 & 5.17 & \textbf{0.72} & \textbf{1.29} & \textbf{0.59} & \textbf{1.01} & \textbf{0.50} & \textbf{1.00} \\
\hline
\hline
Tasks & \multicolumn{8}{c||}{Insertion - 4 rotational symmetries} & \multicolumn{8}{c}{Assembling kits}  \\
\hline
 baseline & 2.21 & 3.55 & 1.33 & 2.41 & 1.25 & 2.37 & 1.25 & 2.38 &  \textbf{8.63} & \textbf{14.15} & 3.91 & 6.96 & 3.85 & 7.23 & 3.40 & 5.93 \\
 di, it4, i-crop & \textbf{1.34} & \textbf{3.02} & \textbf{0.66} & \textbf{1.42} & \textbf{0.55} & \textbf{1.18} & 0.56 & 1.11 &  9.01 & 15.30 & 3.61 & 6.61 & 3.83 & 6.54 & 2.62 & 4.52 \\
 di, it4, f-crop & 2.39 & 4.97 & 0.77 & 1.43 & 0.74 & 1.25 & \textbf{0.53} & \textbf{0.81} & 12.29 & 21.92 & \textbf{2.63} & \textbf{4.85} & \textbf{2.61} & \textbf{4.96} & \textbf{1.84} & \textbf{3.08} \\
\hline
\hline

 & \multicolumn{16}{c}{\textbf{Transport error}} \\
\hline
Tasks & \multicolumn{8}{c||}{Insertion - no rotational symmetry} & \multicolumn{8}{c}{Insertion - 2 rotational symmetries}  \\
\hline
   baseline        & 9.79 & 22.59 & \underline{3.59} & 11.59 & 2.00 & 2.72 & 2.01 & 2.71 & 4.03 &  7.08 & 2.13 & 3.37 & 1.36 & 2.51 & 1.42 & 2.54 \\
 ex, it4, c-rot    & \textbf{8.51} & \textbf{20.26} & 5.53 &  \underline{6.15} & \underline{1.18} & 2.26 & 2.20 & 1.84 & \textbf{2.85} &  \textbf{6.27} & \textbf{1.22} & \textbf{2.36} & 0.81 & 1.34 & 0.66 & 1.22 \\
ex, it4, q-rot     & 20.63 & 36.40 & 6.63 &  7.68 & 3.29 & \underline{1.35} & \textbf{0.82 }& \textbf{1.31} & 8.24 & 14.03 & 1.93 & 3.57 & \textbf{0.67} & \textbf{1.09} & \textbf{0.57} & \textbf{1.10} \\
\hline
\hline
Tasks & \multicolumn{8}{c||}{Insertion - 4 rotational symmetries} & \multicolumn{8}{c}{Assembling kits}  \\
\hline
   baseline       & 7.13 & 15.40 & 1.56 & 3.09 & 1.30 & 2.45 & 1.30 & 2.33 & 6.40 &  9.09 & 3.45 & 6.99 & 2.57 & 4.28 & 2.39 & 4.43 \\
  ex, it4, c-rot  & \textbf{6.63} & \textbf{10.81} & \textbf{1.14} & \textbf{1.79} & 0.67 & \underline{1.00} & \textbf{0.54} & \textbf{0.86} & \textbf{5.33} &  \textbf{8.80} & 3.72 & 6.91 & \textbf{1.88} & \textbf{3.93} & 2.11 & 4.82 \\
 ex, it4, q-rot   & 10.90 & 21.06 & 1.20 & 2.42 & \underline{0.59} & 1.06 & 0.60 & 1.04 & 8.95 & 14.47 & \textbf{3.22} & \textbf{5.84} & 2.65 & 4.68 & \textbf{1.99} & \textbf{3.81} \\
\hline
\end{tabular}
\caption{\label{tab:results}Attention and transport errors: avg. rotation errors (cell left) in deg and avg. translation errors (cell right) in cm.}
\end{table*}

\subsubsection{Metrics} \label{subsubsec:metrics}We measure models' performances with various metrics:
\begin{itemize}
    \item \textbf{Attention error:} theoretical translation and rotation error at picking. For a predicted pose, translation and rotation errors are computed for all grasp-segments. These errors are then filtered to have a translation error less than 1cm and the one with the smallest rotation error is considered as attention error.
    \item \textbf{Transport error:} theoretical translation and rotation error at placing. The place pose is estimated based on a ground truth pick pose. The predicted place pose is then compared to the possible object placement poses, also accounting for the gripper's offset when grasping and for different valid target locations in case multiple identical objects in the assembling kits task. The error is then determined the same way as attention error.
    \item \textbf{Task execution error:} translation and rotation error after simulated execution of the whole pick-and-place process. This is basically the transport error but with an estimated pick pose instead of ground truth pick poses. For this metric, we implement physics simulation based grasping. 
\end{itemize}

\subsubsection{Results}
Due to the innate rotational symmetry of a 2-jaw parallel gripper, we reduce the rotation space for picking to 180°. For discrete training of the attention module, we retain the 10° discretization, thus obtaining 18 rotation bins. For comparability, we use exact training with 17 negative examples. The discrete training of the transport module is identical to TNs with 36 rotation bins (also 10° discretization), and again, for comparability, we use 35 negative examples with exact training. For the initial evaluation, we train models on the 7 block insertion and assembling kits task with different numbers of demonstrations (1, 10, 100, 1000) and store snapshots after different number of weight update steps (1000, 5000, 10000, 20000, 30000). Considering both variations of attention (input crop, feature crop) and transport models (crop rotated, query rotated), in total we have 640 trained attention and 640 trained transport models. We train only one model per hyperparameter configuration per task using the same training and testing datasets. We evaluate each of these models with four different inference methods, regarding the attention error and transport error metrics. For attention models, we use discrete inference with 18 rotation bins, and three iterative inference methods, all of them with 3 iterations and 6 distinct rotations, but with different step scaling factors: 6, 4, 2. Similarly for transport models, discrete inference with 36 rotation bins and three iterative inference methods all with 3 iterations, 12 distinct rotations but with different step scaling factors - 12, 8 and 4 - are evaluated.

After analyzing the models and their performances as described in \ref{subsec:analysis}, as expected the number of training samples and weight update steps are the most influential: with more data and more training leading to better models. 

As for the attention model, the analysis shows, that discrete training combined with iterative inference, in particular with step scaling factor 4 is the best choice. For both rotation and translation precision, feature cropping and thus connecting translation and rotation estimation can provide an additional performance boost if trained with more data or when used in the assembling kits tasks with unknown objects.  Table \ref{tab:results} shows significant improvements when comparing these models to the baseline in terms of average rotation and translation errors of up to 61\% and 68\% respectively. Rotation and translation errors depend on each other and are results of the same model. For better visualization in the tables, underlined entries show if only one of the errors is the lowest while while bold entries mark the overall best performing models.

\begin{table*}[t]
  \centering
  \begin{tabular}{l||rr|rr|rr|rr||rr|rr|rr|rr}
Training samples & \multicolumn{2}{c|}{1} & \multicolumn{2}{c|}{10} & \multicolumn{2}{c|}{100} & \multicolumn{2}{c||}{1000} & \multicolumn{2}{c|}{1} & \multicolumn{2}{c|}{10} & \multicolumn{2}{c|}{100} & \multicolumn{2}{c}{1000} \\
\hline
Tasks & \multicolumn{8}{c||}{Insertion - L-shape} & \multicolumn{8}{c}{Insertion - T-shape}  \\
\hline
  baseline     &36.19 & 16.11 & 13.61 & 5.62 & 6.19 & 1.57 & 9.47 & 3.79 & 67.29 & \underline{11.56} & 36.90 & 12.87 & 17.34 & 5.25 & 14.86 & 6.16 \\
  ic-cr        &\textbf{36.69} & \textbf{14.27} &  8.99 & 3.23 & 4.24 & 0.74 & \underline{1.69} & 0.40 & 78.51 & 14.46 & 30.91 &  9.80 & 17.49 & 4.52 & 12.72 & 2.45 \\
  fc-cr        &38.94 & 16.08 &  \textbf{5.68} & \textbf{0.64} & \textbf{1.61} & \textbf{0.36} & 2.41 & 0.67 & 70.72 & 19.01 & \underline{29.51} & 10.79 & 22.15 & 7.09 & 11.17 & 2.83 \\
  ic-qr        &41.50 & 15.72 &  7.98 & 3.22 & 4.84 & 0.85 & 2.63 & \underline{0.33} & \underline{64.79} & 12.35 & 37.49 &  \underline{9.21} & \textbf{15.15} & \textbf{3.66} &  \textbf{8.42} & \textbf{2.32} \\
  fc-qr        &44.18 & 17.24 &  4.37 & 0.63 & 1.89 & 0.49 & 3.51 & 0.55 & 62.53 & 17.10 & 34.56 & 10.89 & 18.68 & 6.40 & 11.08 & 2.79 \\
\hline
\hline
Tasks & \multicolumn{8}{c||}{Insertion - Cuboid} & \multicolumn{8}{c}{Insertion - Long cuboid}  \\
\hline
  baseline & 9.36 & 0.78 & 3.28 & 0.56 & 3.60 & 0.36 & 2.62 & \underline{0.25} & 8.30 & 0.39 & 3.26 & 0.71 & 2.75 & 0.33 & 2.68 & 0.33 \\
  ic-cr    & 7.28 & 0.41 & 2.10 & 0.39 & 1.24 & 0.31 & 2.06 & 0.35 & 3.07 & \underline{0.35} & 3.37 & 0.39 & 1.26 & 0.32 & 1.60 & 0.45 \\
  fc-cr    & \textbf{6.58} & \textbf{0.36} & \textbf{1.88} & \textbf{0.34} & 1.24 & 0.38 & 2.20 & 0.37 & 2.85 & \underline{0.35} & 5.98 & 2.52 & 1.43 & 0.32 & 1.35 & \underline{0.23} \\
  ic-qr    &38.66 & 0.59 & 5.90 & 0.44 & 1.12 & \underline{0.24} & 1.60 & 0.35 & 2.45 & 0.48 & \textbf{1.71} & \textbf{0.31} & \textbf{1.07} & \textbf{0.30} & 1.24 & 0.47 \\
  fc-qr    &40.52 & 0.59 & 5.89 & 0.45 & \textbf{1.04} & \textbf{0.24} & \underline{1.20} & 0.32 & \underline{2.39} & 0.46 & 5.55 & 2.81 & 1.87 & 0.66 & \underline{1.11} & 0.27 \\
\hline
\hline
Tasks & \multicolumn{8}{c||}{Insertion - Double-L-shape} & \multicolumn{8}{c}{Insertion - Cube}  \\
\hline
 baseline &36.62 & 36.13 & 7.77 & 3.78 & 20.49 & 17.74 & 21.95 & 19.07 &  \textbf{9.16} &  \textbf{0.40} & 3.12 & \underline{0.34} & 2.41 & \underline{0.21} & 2.49 & 0.27 \\
 ic-cr    &41.39 & 36.60 & 7.22 & \underline{3.66} & 12.29 &  8.74 & 18.14 & 14.63 & 16.89 &  9.31 & 2.96 & 0.37 & 1.21 & 0.24 & \underline{1.00} & 0.24 \\
 fc-cr    &29.61 & \underline{24.70} & 9.78 & 5.35 &  9.42 &  \underline{6.51} &  \textbf{4.79} &  \textbf{4.13} & 15.23 & 10.12 & 2.48 & 0.43 & 1.35 & 0.28 & 1.11 & 0.26 \\
 ic-qr    &38.47 & 34.96 & \underline{6.37} & 3.69 & 12.11 &  9.14 & 17.56 & 14.65 & 20.96 &  3.81 & 2.18 & 0.41 & \underline{1.15} & 0.25 & 1.04 & \underline{0.23} \\
 fc-qr    &\underline{27.12} & 21.15 & 9.25 & 5.40 &  \underline{8.77} &  6.91 &  5.03 &  4.25 & 21.19 &  3.13 & \underline{1.74} & 0.43 & 1.19 & 0.28 & 1.13 & 0.26 \\
\hline
\hline
Tasks & \multicolumn{8}{c||}{Insertion - Cross} & \multicolumn{8}{c}{Assembling kits}  \\
\hline
 baseline & \underline{4.28} & 12.05 & 3.92 & 1.56 & 9.02 & 9.00 & 3.28 & 1.64 & 21.23 & 24.48 & 16.18 & 8.41 & \underline{11.57} & 3.70 & 12.79 & 3.98 \\
 ic-cr    &12.10 & 19.50 & \textbf{3.19} & \textbf{1.07} & 6.91 & 8.74 & \textbf{2.40} & \textbf{1.37} & 21.18 & 24.71 & 16.11 & 7.42 & 12.58 & 4.53 &  \underline{9.48} & 3.40 \\
 fc-cr    & 9.35 & 12.36 & 3.51 & 2.31 & 2.05 & \underline{1.50} & 5.69 & 7.68 & 20.35 & 24.42 & 15.09 & 5.21 & 12.70 & 3.53 & 14.03 & 9.34 \\
 ic-qr    &23.57 & 13.76 & 4.40 & 1.72 & 6.78 & 8.40 & 2.43 & 1.42 & 22.37 & 25.30 & 17.23 & 6.66 & 13.85 & 5.00 & 10.93 & \underline{3.26}  \\
 fc-qr    &23.54 &  \underline{4.70} & 4.57 & 3.13 & \underline{1.99} & 1.57 & 7.03 & 8.38 & \textbf{19.33} & \textbf{24.03} & \textbf{11.51} & \textbf{4.45} & 12.26 & \underline{3.21} & 11.66 & 9.45 \\
\hline
\end{tabular}
\caption{\label{tab:transporter-res}Task execution error: avg. rotation errors (cell left) in deg and avg. translation errors (cell right) in cm}
\end{table*}

The transport module profited from both exact training and iterative inference, especially with step scaling factor 4. The query rotated variant provided similar results as the input rotated variant but at lower computational costs. Table \ref{tab:results} summarizes the transport errors on different task groups and shows up to 60\% improvement in average rotation error and up to 63\% improvement in average translation error when using the proposed training and inference methods.

 Our analysis highlights two attention and two transport configurations as most promising. Both attention models are trained with discrete rotations and evaluated with iterative rotations with scaling step size 4. The only difference is whether input crop or feature crop is used for rotation estimation. Similarly, the training and inference configuration of the transport modules are identical (exact training with iterative inference with step scaling 4), they only differ in their model variants: crop rotated or query rotated. In the following we combine these modules into 4 complete transporter models for further analysis. We refer to these only by the combination of their respective module variants, for example ic-cr for input cropped attention combined with crop rotated transport. As baseline we use the combination of an input cropped attention and a crop rotated transport module, both with discrete training and discrete inference. Table \ref{tab:transporter-res} shows the task execution errors of the combined transporter models for all tasks. All models struggle with more complex objects (objects with more knicks: T-shape, double-L-shape, cross). In case of simpler objects, our proposed methods consistently outperform the baseline in terms of rotation errors and have similar translation errors. Additionally in these cases, models trained with 100 demonstrations perform as good as models trained with 1000 demonstrations.

\subsection{Real Robot Experiments}
Our real world setup consists of an UR10e robot equipped with a Robotiq 2F-140 gripper and an Intel RealSense d415 camera. We train and test the best performing model combination from the simulated experiments against the baseline using the interactive teaching method described below. 
\subsubsection{Interactive Teaching}
During interactive teaching, a randomly initialized model is trained and evaluated simultaneously.
First, the operator sets up a task and the robot is prompted to record an observation and infer a pick pose.
Then, the robot moves to a pre-pick pose and the operator can correct the predicted pose via cartesian servoing. 
After grasping the object a place pose is computed based on the corrected pick pose. 
Analogous to picking, the robot moves to the estimated pre-place pose and the operator can correct this pose again. The observation and the corrected poses form a training sample and are stored in the training dataset. After each demonstration, the dataset is expanded with one sample and the model is trained for 500 weight update steps (approximately 2 minutes). During this period, the operator sets up a new task instance for the next demonstration.



\subsubsection{Results}
We train the models once to solve a pick-and-place task, where the robot has to pick up a cube and place it into a quadratic fixture (Fig. \ref{fig:transport-real}). We perform 10 demonstrations while teaching each model, resulting in 5000 weight update steps. This takes approximately 25 minutes when using a query rotated transport module and ca. 40 minutes with a crop rotated variant. After teaching, we test the newly trained model on 20 randomly set up task configurations. A task execution counts as successful, when the cube lies completely flat on the workspace inside the fixture. Table \ref{tab:real_robot_exp} summarizes the results.

\begin{table}[!]
\centering
    \begin{tabular}{c||c|c|c|c|c}
        model & baseline & ic-cr & fc-cr & ic-qr & fc-qr \\ 
        \hline
        success rate  & 60\% & 70\% & 40\% & 80\% & 95\% \\
        \hline
    \end{tabular}
    \caption{Success rates for real world experiment}
    \label{tab:real_robot_exp}
\end{table}

\section{Conclusion}
This paper aims at improving precision in pick-and-place tasks. To this end, we define the exact training method and the iterative inference method for Transporter Networks as well as the architectural changes: feature cropping in the attention (pick) module and query rotating in the transport (place) module. We systematically analyse the impact of these changes on models' precision (translation and rotation errors) on simulated tasks. We find, that the attention module is positively affected by iterative inference, and the transport module by both, exact training and iterative inference. Feature cropping improved performance in some tasks, while query rotating retained performance at lower computational costs. When combining the models to a complete pick-and-place pipeline we find, that our methods significantly lower rotation errors when compared to the baseline with slight improvements in terms of translation errors. We also find, that all models still struggle when handling more complex shapes, especially with low numbers of training data.

With the goal of precise object handling, we have to keep in mind that it can not be solved by pick-and-place only. 
It requires some other types of object manipulation that are dynamically constrained like pushing, aligning or inserting, which on the other hand require continuous feedback. 
We look forward to exploring and including these directions into our work in the future.

\bibliographystyle{IEEEtran}
\bibliography{references}
\end{document}